\documentclass{article}

\usepackage{microtype}
\usepackage{graphicx}
\DeclareUnicodeCharacter{2212}{-}
\usepackage{xcolor}
\usepackage{amsmath}
\usepackage{pbox}
\usepackage{amssymb}
\usepackage{array,makecell}
\usepackage{booktabs} % for professional tables
\usepackage{subcaption}

\usepackage{hyperref}

\usepackage[accepted]{icml2021}

\icmltitlerunning{When Differential Privacy Meets Interpretability: A Case Study}
\begin{document}

\twocolumn[
\icmltitle{When Differential Privacy Meets Interpretability: A Case Study}
% \\Case Studies in Medical Imaging}



\icmlsetsymbol{equal}{*}

\begin{icmlauthorlist}
\icmlauthor{Rakshit Naidu}{equal,man,cmu,om}
\icmlauthor{Aman Priyanshu}{equal,man}
\icmlauthor{Aadith Kumar}{man,upenn}
\icmlauthor{Sasikanth Kotti}{om,iitj}
\icmlauthor{Haofan Wang}{cmu,om}
\icmlauthor{Fatemehsadat Mireshghallah}{ucsd,om}
% \icmlauthor{Fiuea Rrrr}{to}
% \icmlauthor{Tateu H.~Yasehe}{ed,to,goo}
% \icmlauthor{Aaoeu Iasoh}{goo}
% \icmlauthor{Buiui Eueu}{ed}
% \icmlauthor{Aeuia Zzzz}{ed}
% \icmlauthor{Bieea C.~Yyyy}{to,goo}
% \icmlauthor{Teoau Xxxx}{ed}
% \icmlauthor{Eee Pppp}{ed}
\end{icmlauthorlist}

\icmlaffiliation{man}{Manipal Institute of Technology}
\icmlaffiliation{cmu}{Carnegie Mellon University}
\icmlaffiliation{om}{OpenMined}
\icmlaffiliation{upenn}{University of Pennsylvania}
\icmlaffiliation{iitj}{IIT Jodhpur}
\icmlaffiliation{ucsd}{University of California, San Diego}

\icmlcorrespondingauthor{Rakshit Naidu}{rnemakal@andrew.cmu.edu}
% \icmlcorrespondingauthor{Eee Pppp}{ep@eden.co.uk}

\icmlkeywords{Machine Learning, ICML}

\vskip 0.3in
]



\printAffiliationsAndNotice{\icmlEqualContribution} % otherwise use the standard text.

\begin{abstract}
Given the increase in the use of personal data for training Deep Neural Networks (DNNs) in tasks such as medical imaging and diagnosis, differentially private training of DNNs is surging in importance and there is a large body of work focusing on providing better privacy-utility trade-off. However, little attention is given to the interpretability of these models, and how the application of DP  affects the quality of interpretations. We propose an extensive study into the effects of DP training on DNNs, especially on medical imaging applications, on the APTOS dataset.
\end{abstract}

\section{Introduction}

The application and development of Machine Learning (ML) in the field of medicine and health has grown exponentially. 
With the recent advances  in employing AI for health, one can  see the great potential it holds \cite{priyanshu2021fedpandemic, kaissis2021end}.
However, healthcare data contains sensitive information which must be handled under security protocols, which protect subject privacy. At the same time, model results and predictions must be interpretable allowing medical experts involved to study and validate the evaluations~\cite{suriyakumar2021chasing,icip2021}. This clearly identifies a problem within computer vision, where both accountability and privacy must be addressed for certain use-cases.
DP~\cite{DworkRoth14, Abadi_2016} is defined as an extensive tool that constitutes strong privacy guarantees for algorithms on a given data distribution by providing the overall patterns within the dataset while withholding information about individuals. Interpretable Machine learning is defined as a collection of algorithmic implements, which allow the user to understand how the model arrived at a specific decision. These two technologies, if correctly leveraged can further refine the present-day implementations and make them more practical, adding dimensions of trust, corrective feedback and contestability of claim to  black-box models.
%make medical imaging tasks commercially viable while upholding patient privacy.
%
%A fusion of Differential Privacy (DP) with interpretable machine learning can further refine the present-day implementations and make them more practical. Making these models not only privacy-preserving but also understandable

% Differential Privacy (DP) is an extensive tool that constitutes strong privacy guarantees for algorithms on a given dataset by describing the patterns of groups within the dataset while withholding information about individuals in the dataset. 

\begin{figure}[t]
    \centering
        \begin{subfigure}{0.4\textwidth}
     \includegraphics[width=\linewidth]{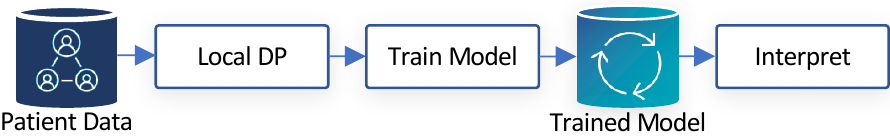}
    \footnotesize 
     \caption{ Local Differential Privacy}
     \label{fig:mnistablation}
    \end{subfigure}
    
    \begin{subfigure}{0.4\textwidth}
     \includegraphics[width=\linewidth]{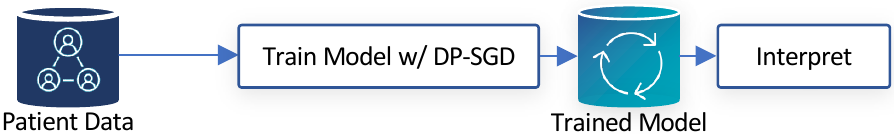}
     \footnotesize
     \caption{Global Differential Privacy with DP-SGD}
     \label{fig:svhnablation}
    \end{subfigure}
    % \begin{subfigure}{0.28\textwidth}
    %  \includegraphics[width=\linewidth]{figs/mnist_dpsgd_15.png}
    %  \caption{MNIST DP-SGD for $\varepsilon=15$}
    %  \label{fig:blogs-2dom-unc}
    % \end{subfigure}
    \caption{We benchmark two application schemes of differential privacy: (a) local DP where noise is directly added to the data samples and (b) global DP where the raw data is collected and the noise is added during the training/analysis in a centralized manner.}
    \label{fig:DPmodes}
    \vspace{-2ex}
\end{figure}

%
There are different models of applying differential privacy, based on where the ``privacy barrier'' is set, and after which stage in the pipeline we need to provide privacy guarantees~\cite{mirshghallah2020privacy, bebensee2019local}, as shown in Figure~\ref{fig:DPmodes}. (1) Local DP is comprised of applying noise directly to the user data. This way, the data itself can be shared with untrusted parties and the leakage is bounded by the privacy budget $\varepsilon$. (2) Global DP, which is based on the assumption that there exists a trusted party, a centralized server, which collects the data, and then applies differentially private learning algorithms on the collected data  to produce  models or analysis with bounded information leakage. A prominent one of such algorithms is DP-SGD~\cite{Abadi_2016, chaudhuri2011differentially}, which consists of clipping gradients and adding noise at each step of the optimization. 
Due to this addition of noise, interpreting differentially private models is more difficult than conventionally trained ones~\cite{ModelExpDPShokri20}.

In this paper, we set out to explore this problem of interpreting differentially private models, in health use-cases, more extensively. We provide the first benchmark of exploring interpretability, specifically through class activation maps, in DNNs trained using DP. We train DP-DNNs with a wide range of privacy budgets in both local and global DP settings, to study the effects they have on model interpretability. We utilize Grad-CAM \cite{GradCAM16} as our interpretability method and use the Cats vs Dogs and APTOS~\cite{APTOS19} datasets, to train our models in both general and medical setups. 
We show how different levels of privacy budget ($\varepsilon$) effect the interpretablity and utility of the model, and explore this three dimensional trade-off space for local vs. global application of DP.
% The noise used is sampled from a Gaussian Distribution $G(0, \sigma^{2})$.
%%%%%%%%%%%%%%%%%

%%%%%%%%%%%%%%%
\section{Related Work}

\subsection{Differential Privacy}
\textit{Definition 1}: Given a randomized mechanism $\mathcal{A}$ and any two neighboring datasets $D_{1}, D_{2}$ (\emph{i.e.} they differ by a single individual data element), $\mathcal{A}$ is said to be $(\varepsilon, \delta)$-differentially private for any subset $S \subseteq \mathcal{R}$ (where $\mathcal{R}$ denotes the range).
% \footnote{In this work, we exclusively use Gaussian noise}.
\begin{equation}
\Pr\left[ \mathcal{A}\left( D_{1}\right) \in S\right] \leq e^{\varepsilon }\cdot\Pr\left[ \mathcal{A}\left( D_{2}\right) \in S\right] + \delta
\end{equation}
Here, $\epsilon \geq 0, \delta \geq 0$. A $\delta = 0$ case corresponds to pure differential privacy, while both $\epsilon = 0, \delta = 0$ leads to an infinitely high privacy domain. Finally, $\epsilon = \infty$ provides no privacy (non-DP) guarantees.

% {\rakshit{I think we should include more details about DP and what Gradient (DP-SGD) and Local-DP mean and why one could use these methods.}}

The privacy in DP models can be quantified with parameters such as epsilon ($\varepsilon$) and delta ($\delta$). DP-SGD \cite{Abadi_2016} adds noise to the gradients at each step during training using a clipping factor ($S$) and noise multiplier ($z$). The amount of noise added to the model can be linked to to the privacy level that the model can achieve. Theoretically, a lower value of $\epsilon$ signifies a higher privacy level and this increased privacy degree is understandably, achieved at the expense of model performance due to the addition of the noise. The practical implication of this, however, includes a direct impact on the model performance and hence, its interpretability. Reduced privacy requirements allow the addition of noise where models can achieve appropriate performance without any significant resource expense. On the other hand, high privacy standards add a significantly large magnitude of noise which may lead to a decrease in the model  interpretations. Further, noise addition may even lead to the non-convergence of some models in the worst case. In the scope of this paper, we take $\delta = 10^{-4}$ and experiment on various privacy levels $\epsilon = 0.5,1,5,10$.

\subsection{Interpretability}

In recent years, Convolutional Neural Networks (CNNs) have been involved in progressing major vision tasks such as image captioning~\cite{8803108}, image classification~\cite{classification19} and semantic segmentation~\cite{SemanticSeg18}, to name a few. Despite these advancements, CNNs are treated as black-box architectures when applied to sensitive environments. 

To solve explainability measures in CNNs, Class Activation Mappings (CAMs) was proposed for interpreting image classification tasks by \citeauthor{zhou2015learning}. CAMs refer to the linear combination of the weights and the activation maps produced by the Global Average Pooling (GAP) layer. In the scope of this paper, we utilize Grad-CAM \cite{GradCAM16} for producing explanation maps. Grad-CAM is denoted by : 

\begin{equation}
    L^c_{Grad−CAM}=ReLU\left(\sum\limits_{k} \alpha_{c}^{k} A^{k}\right)
\end{equation}
where $\alpha_c^k$ represents the neuron importance weights. $\alpha_c^k=\frac{1}{Z}\sum\limits_{i}\sum\limits_{j}\frac{\partial Y_c}{\partial A^k_{i j}}$ where $Y_c$ is the score computed for the target class, $(i,j)$ represents the location of the pixel and $Z$ denotes the total number of pixels.

\section{Evaluations}

% DPSGD on APTOS (Resnet and Alexnet) and Cats and Dogs (Resnet and VGG-16)

% Case Study 1: Local DP on APTOS (Resnet and Alexnet)

% Case Study 2: Laplacian noise with Cats and Dogs (Resnet and VGG-16) [Test Accuracy]

We evaluate on two datasets : APTOS and the Cats vs Dogs dataset. We use DP-SGD with a learning rate of 0.0001 and a batch size of 32 over the Cats vs Dogs dataset for 10 epochs. We utilize the APTOS Local DP (LDP) datasets as demonstrated by \citeauthor{BenchmarkDP20}.

\begin{figure}[!htb]
     \centering
     \begin{subfigure}[h]{0.5\textwidth}
         \centering
         \includegraphics[width=\textwidth]{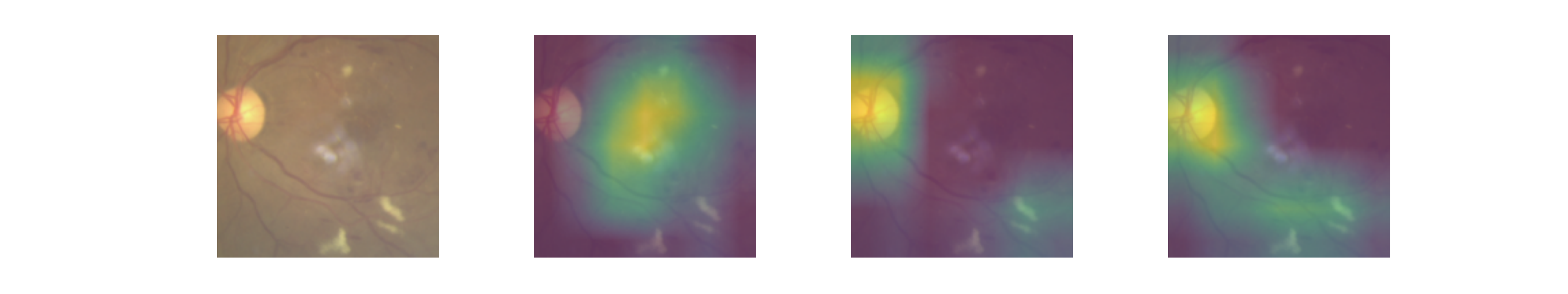}
         \caption{ResNet18}
         \label{fig:resnet_ldp_aptos}
     \end{subfigure}
     \begin{subfigure}[h]{0.5\textwidth}
         \centering
         \includegraphics[width=\textwidth]{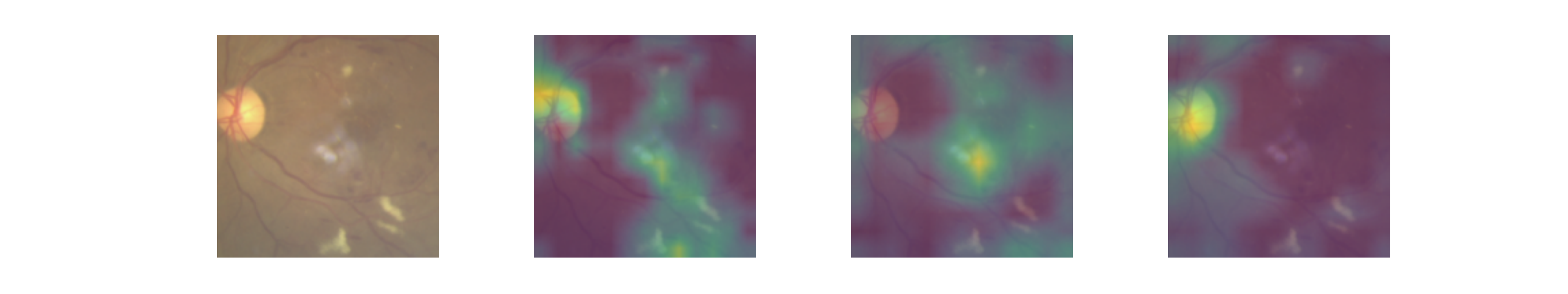}
         \caption{AlexNet}
         \label{fig:alexnet_ldp_aptos}
     \end{subfigure}
     \caption{Grad-CAM Explanation maps on Local DP for APTOS dataset with Input image, $\epsilon = 0.25, 0.5, 1$ (left to right).
    %  \red{Provide some analysis, what does this figure shows?}
    }
     \label{fig:ldp_aptos}
\end{figure}

In Fig~\ref{fig:ldp_aptos}, we observe that as $\epsilon$ increases (degree of privacy decreases), we approach explanations with better quality.  

\begin{figure}[!htb]
\small
\vskip 0.2in
\begin{center}
\centerline{\includegraphics[width=0.8\linewidth]{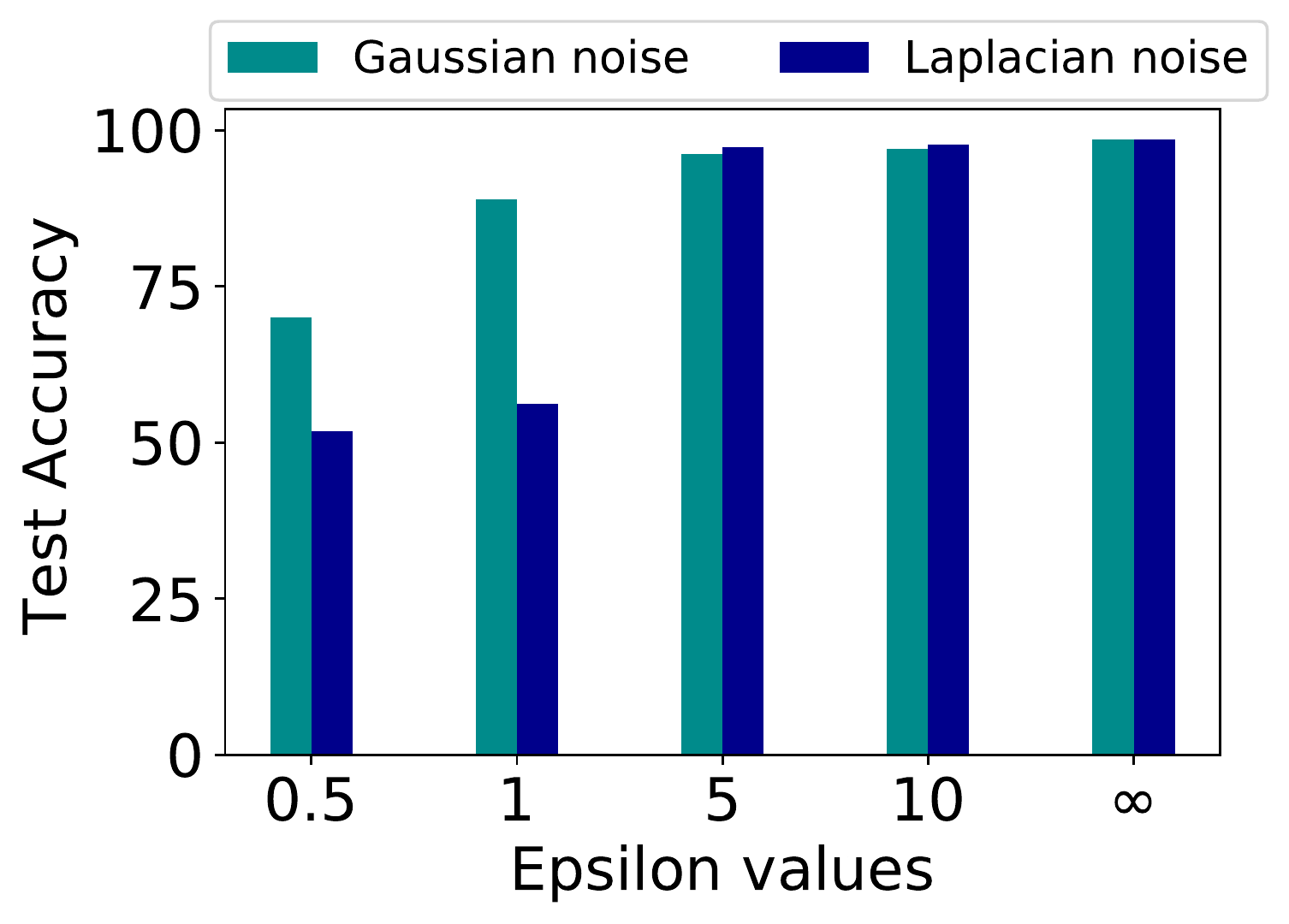}}
\caption{Comparison of Test accuracies on Cats vs Dogs with noise added in DP-SGD.}
\label{fig:testacc_catsdogs}
\end{center}
\vskip -0.2in
\end{figure}

It is well known that Gaussian distribution follows approximate DP~\cite{balle2018improving} while the Laplacian distribution (with a parameter of $\dfrac{1}{\epsilon}$) satisfies $\epsilon$-DP~\cite{DworkRoth14}. Fig~\ref{fig:testacc_catsdogs} shows test accuracies on both Gaussian and Laplacian noise distributions on DP-SGD. We notice that the gap between $\epsilon = 1$ and $\epsilon = 5$ is quite large (almost 25\%) which can be explained by the decrease in variance or ``spread'' over which the noisy values are sampled. 

\begin{table*}[h]
\centering
\begin{tabular}{llllll}
\hline
Metrics     & $\epsilon = 0.5$ & $\epsilon = 1$ & $\epsilon = 5$ & $\epsilon = 10$ & $\epsilon = \infty$ \\ \hline
Average Drop (lower the better) \%                   & 13.29            & 11.66          & 6.76       & 5.81        & 2.05            \\
Average Inc (higher the better) \%                    & 67.52            & 62.50          & 84.96      & 86.8        & 95.04           \\
% Test Accuracy (higher the better) \%                  & 69.98            & 88.96          & 96.24      & 97.07       & 98.5            \\
Average normalized input scores (higher the better) \% & 75.27            & 79.10          & 87.88      & 89.40       & 97.22           \\ \hline
\end{tabular}
\caption{Results on VGG-16 network on the Cats vs Dogs dataset with Gaussian noise (approximate DP) in DP-SGD.}
\label{gaussian_vgg_catdog}
\end{table*}

\begin{table*}[h]
\centering
\begin{tabular}{llllll}
\hline
Metrics     & $\epsilon = 0.5$ & $\epsilon = 1$ & $\epsilon = 5$ & $\epsilon = 10$ & $\epsilon = \infty$ \\ \hline
Average Drop (lower the better) \%                   & --            & 9.61          & 4.39       & 4.18        & 2.05            \\
Average Inc (higher the better) \%                    & --            & 66.84          & 90.96      & 92.44        & 95.04           \\
% Test Accuracy (higher the better) \%                  & 69.98            & 88.96          & 96.24      & 97.07       & 98.5            \\
Average normalized input scores (higher the better) \% & --            & 86.48          & 91.78      & 92.61       & 97.22           \\ \hline
\end{tabular}
\caption{Results on VGG-16 network on the Cats vs Dogs dataset with Laplacian noise (pure DP) in DP-SGD. Note that $\epsilon = 0.5$ results are not displayed as the model here is too private to infer relevant features.} 
\label{laplacian_vgg_catdog}
\end{table*}

We evaluate interpretability on DP-models using two metrics that were introduced by \citeauthor{Chattopadhay_2018} :

\begin{itemize}
    \item \textit{Average Drop \%}: The Average Drop refers to the maximum positive difference in the predictions made by the prediction using the input image and the prediction using the saliency map. It is given as: $\sum_{i=1}^{N} \frac{max(0, Y_{i}^{c} - O_{i}^{c})}{Y_{i}^{c}} \times 100$. Here, \textit{Y\textsubscript{i}\textsuperscript{c}} refers to the prediction score on class \textit{c} using the input image \textit{i} and \textit{O\textsubscript{i}\textsuperscript{c}} refers to the prediction score on class \textit{c} using the saliency map produced over the input image \textit{i}.

    \item \textit{Increase in Confidence \%}: The Average Increase in Confidence is denoted as: $\sum_{i=1}^{N} \frac{Fun(Y_{i}^{c} < O_{i}^{c})}{N} \times 100$ where \textit{Fun} refers to a boolean function which returns 1 if the condition inside the brackets is true, else the function returns 0. The symbols are referred to as shown in the above experiment for Average Drop.
\end{itemize}

\begin{figure}[!htb]
     \centering
     \begin{subfigure}[h]{0.3\textwidth}
         \centering
         \includegraphics[width=\textwidth]{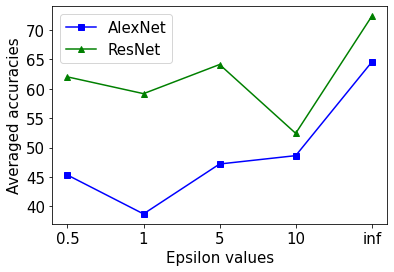}
         \caption{Different $\varepsilon$ Values}
         \label{fig:epsmasks}
     \end{subfigure}
     \begin{subfigure}[h]{0.3\textwidth}
         \centering
         \includegraphics[width=\textwidth]{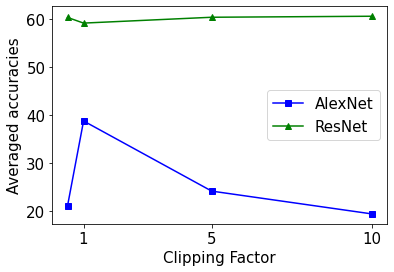}
         \caption{Different Clipping Factors}
         \label{fig:clipmasks}
     \end{subfigure}
     \caption{Averaged accuracies of inputs masked with their explanations over AlexNet and ResNet networks for (a) different $\varepsilon$ values and (b) different clipping factors (S) with $\varepsilon=1$. We quantitatively show that there's a significant gap between privacy and explanation quality.
    %  \red{Provide some analysis, what does this figure shows?}
    }
     \label{fig:avgacc_inpmasks}
\end{figure}

Tables~\ref{gaussian_vgg_catdog} and~\ref{laplacian_vgg_catdog} showcase results of these metrics on the Cats vs Dogs dataset. We notice a consistent decrease with Average Drop and steady increase (except in the case of $\epsilon = 1$) with Average Increase in Confidence experiments. The final metric displayed is the average scores on the point-wise multiplied input with the explanation maps which increases with the value of $\epsilon$, as expected. The same metric is graphically shown in Fig~\ref{fig:epsmasks} for DP-SGD with APTOS.

\begin{figure}[!htb]
     \centering
     \begin{subfigure}[h]{0.3\textwidth}
         \centering
         \includegraphics[width=\textwidth]{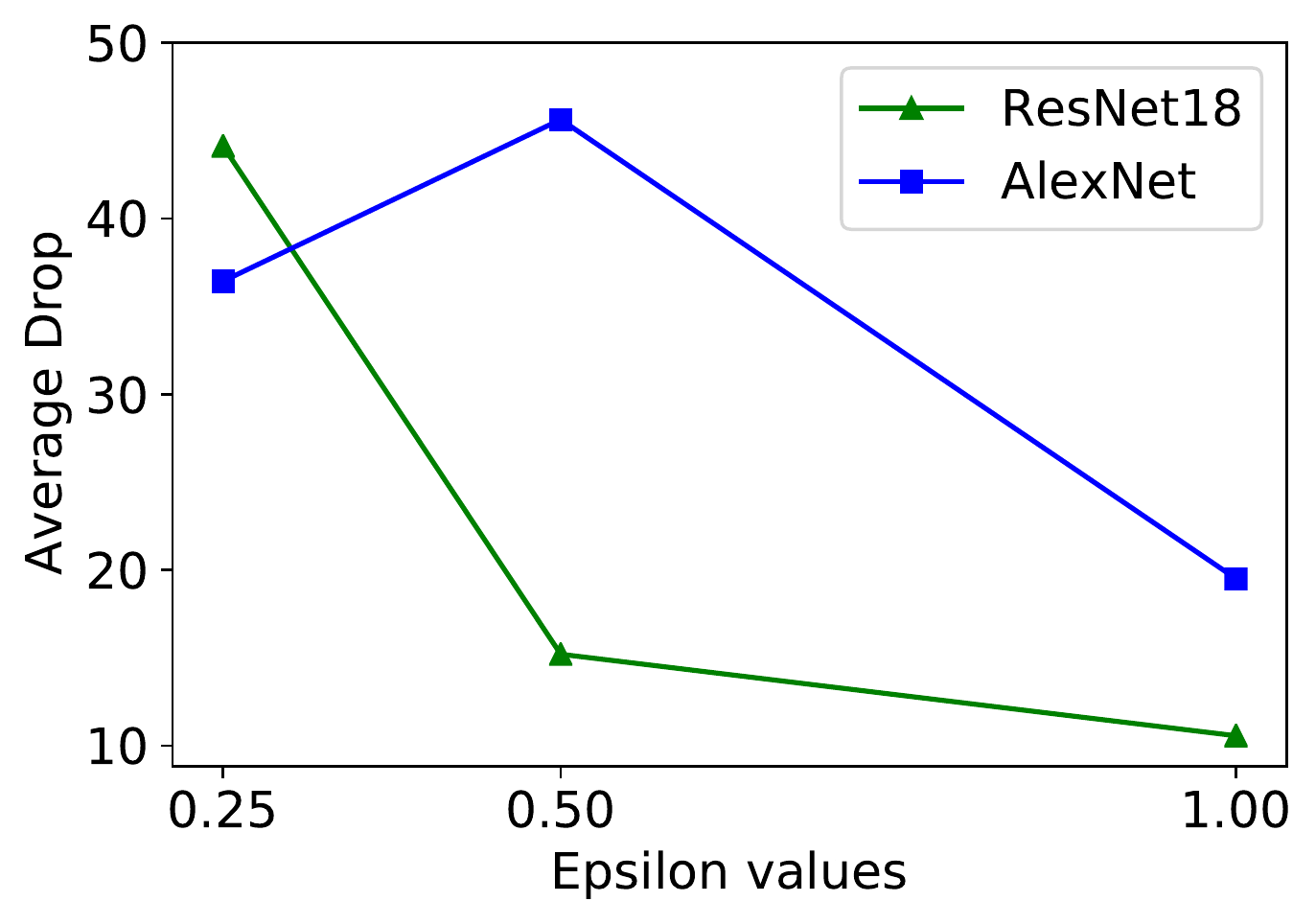}
         \caption{Average Drop (\%)}
         \label{fig:avgdrop_ldp}
     \end{subfigure}
     \begin{subfigure}[h]{0.3\textwidth}
         \centering
         \includegraphics[width=\textwidth]{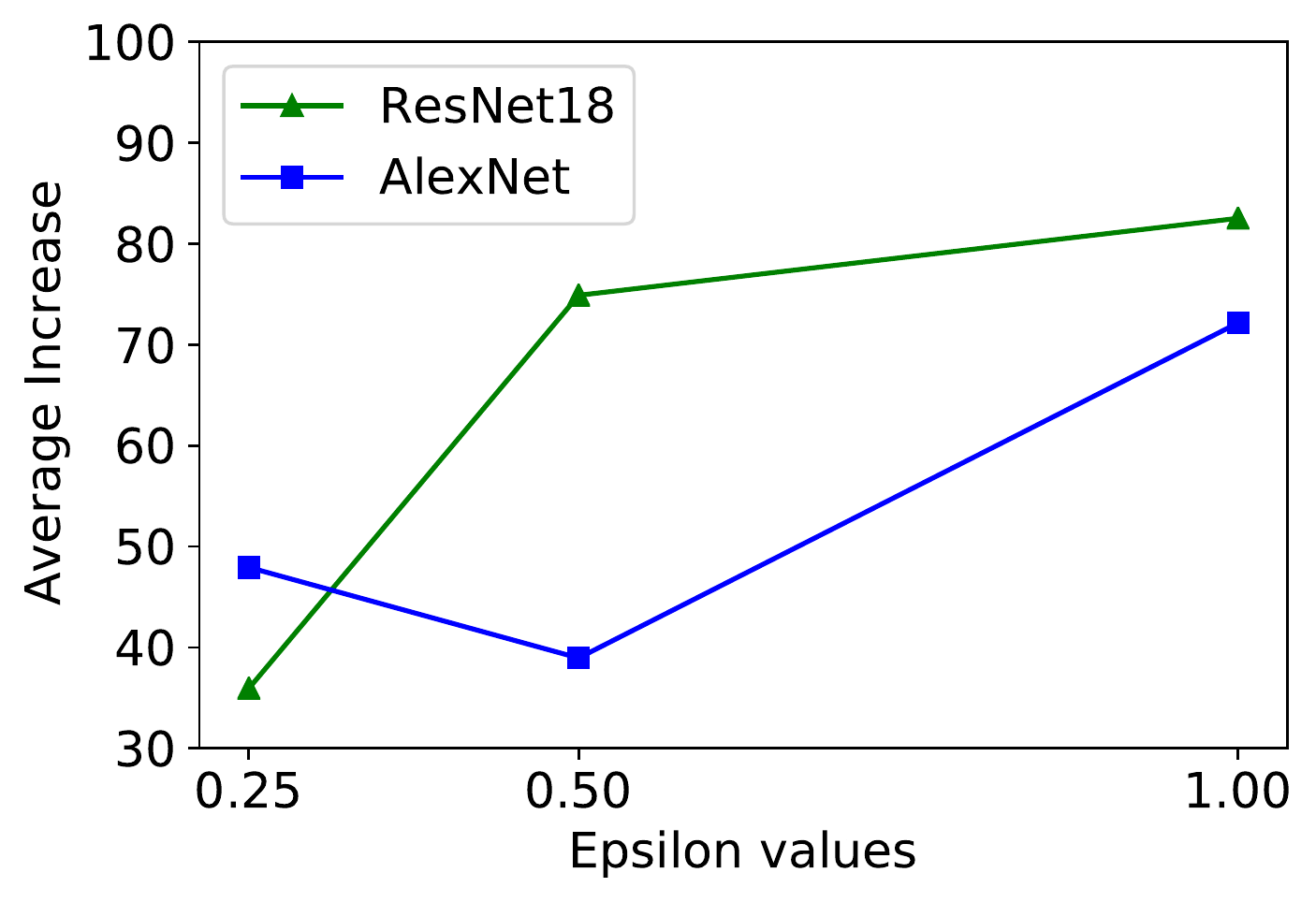}
         \caption{Average Increase (\%)}
         \label{fig:avginc_ldp}
     \end{subfigure}
     \caption{Average Drop and Average Increase in Confidence scores on the APTOS dataset for Local DP. We notice that $\epsilon = 0.5$ for AlexNet opposes the expected trend, probably due to the size of the network as AlexNet is a much smaller network when compared to ResNet18. Recent work by \citeauthor{DiffPrivFlorianICLR21} shows that models with lower parameters (smaller-sized models) exhibit significant changes in accuracy when varying levels of DP noise is added.   
    %  \red{Provide some analysis, what does this figure shows?}
    }
     \label{fig:avgdropsandinc_ldp}
\end{figure}

ResNet18 \cite{BenchmarkDP20} and AlexNet \footnote{We train them using DP-SGD with $0.01$ lr and $25$ epochs.} are adopted in our work over the APTOS dataset. We compute the masked input via point-wise multiplication between the saliency map and the input, and calculate the average output scores of these masked inputs over the testing set. In Fig~\ref{fig:clipmasks}, we observe the effect of $S$ (clipping factor) on the output scores. As $S$\footnote{$S$ $\propto$ $\sigma$} increases (after $S=1$), we notice a drastic decrease in accuracy with AlexNet. This result is quite intuitive as AlexNet is a much smaller network than the ResNet. Therefore, the effect of DP noise is more evident in AlexNet as it has $\thicksim$2M parameters while ResNet has $\thicksim$11M parameters \cite{DiffPrivFlorianICLR21}. We also notice significant changes in the trends observed with Local DP as well (shown in~\ref{fig:avgdrop_ldp} and~\ref{fig:avginc_ldp}) with AlexNet probably due to the above reason.
In Fig~\ref{fig:epsmasks}, we notice a drop in accuracies in $\varepsilon=1$ and a steep increase (nearly 15\%) between $\varepsilon=10$ and $\varepsilon=\infty$. This increase is attributed to the gap between having some privacy to no privacy. 

\begin{figure}[!htb]
     \centering
     \begin{subfigure}[h]{0.5\textwidth}
         \centering
         \includegraphics[width=\textwidth]{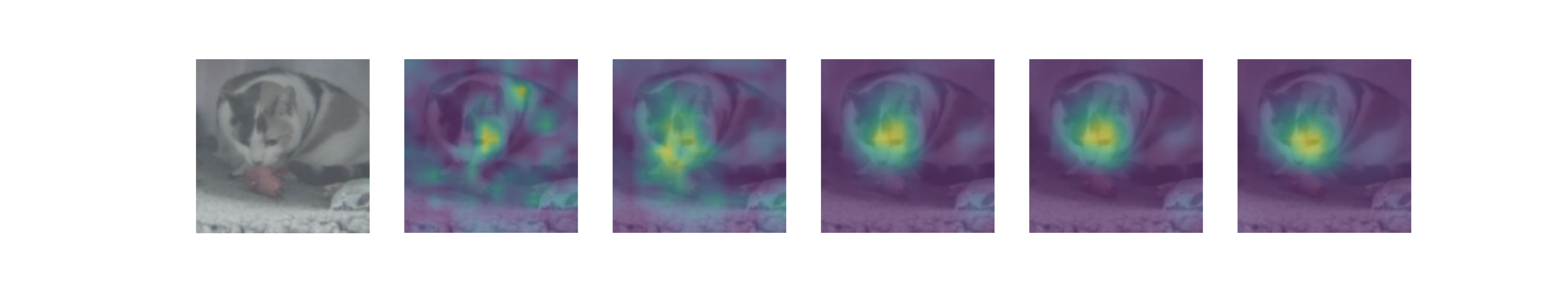}
         \includegraphics[width=\textwidth]{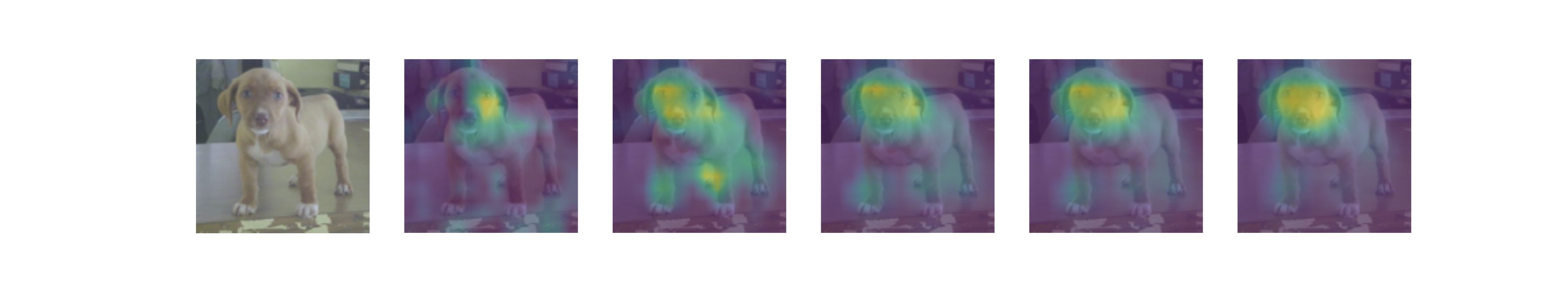}
         \caption{Gaussian DP-SGD}
         \label{fig:gauss_dpsgd}
     \end{subfigure}
     \begin{subfigure}[h]{0.5\textwidth}
         \centering
         \includegraphics[width=\textwidth]{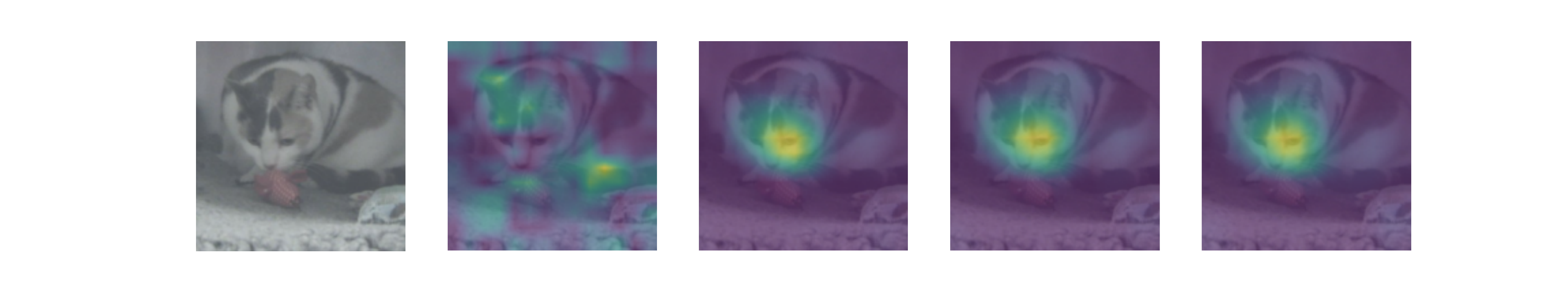}
         \includegraphics[width=\textwidth]{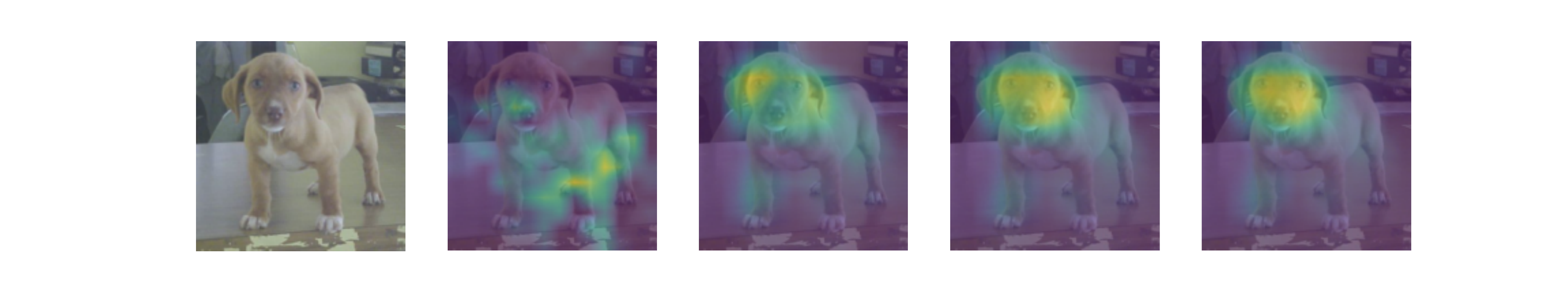}
         \caption{Laplacian DP-SGD}
         \label{fig:laplace_dpsgd}
     \end{subfigure}
     \caption{Grad-CAM Explanation maps on ``Cat'' and ``Dog'' for Input image, $\epsilon = 0.5, 1, 5, 10, \infty$ (left to right). Note that $\epsilon = 0.5$ map for Laplacian variant is not displayed as the model here is too private to infer any relevant features.
    %  \red{Provide some analysis, what does this figure shows?}
    }
     \label{fig:catdogexplanation}
\end{figure}

\begin{figure}[!htb]
\small
\vskip 0.2in
\begin{center}
\centerline{\includegraphics[width=0.8\linewidth]{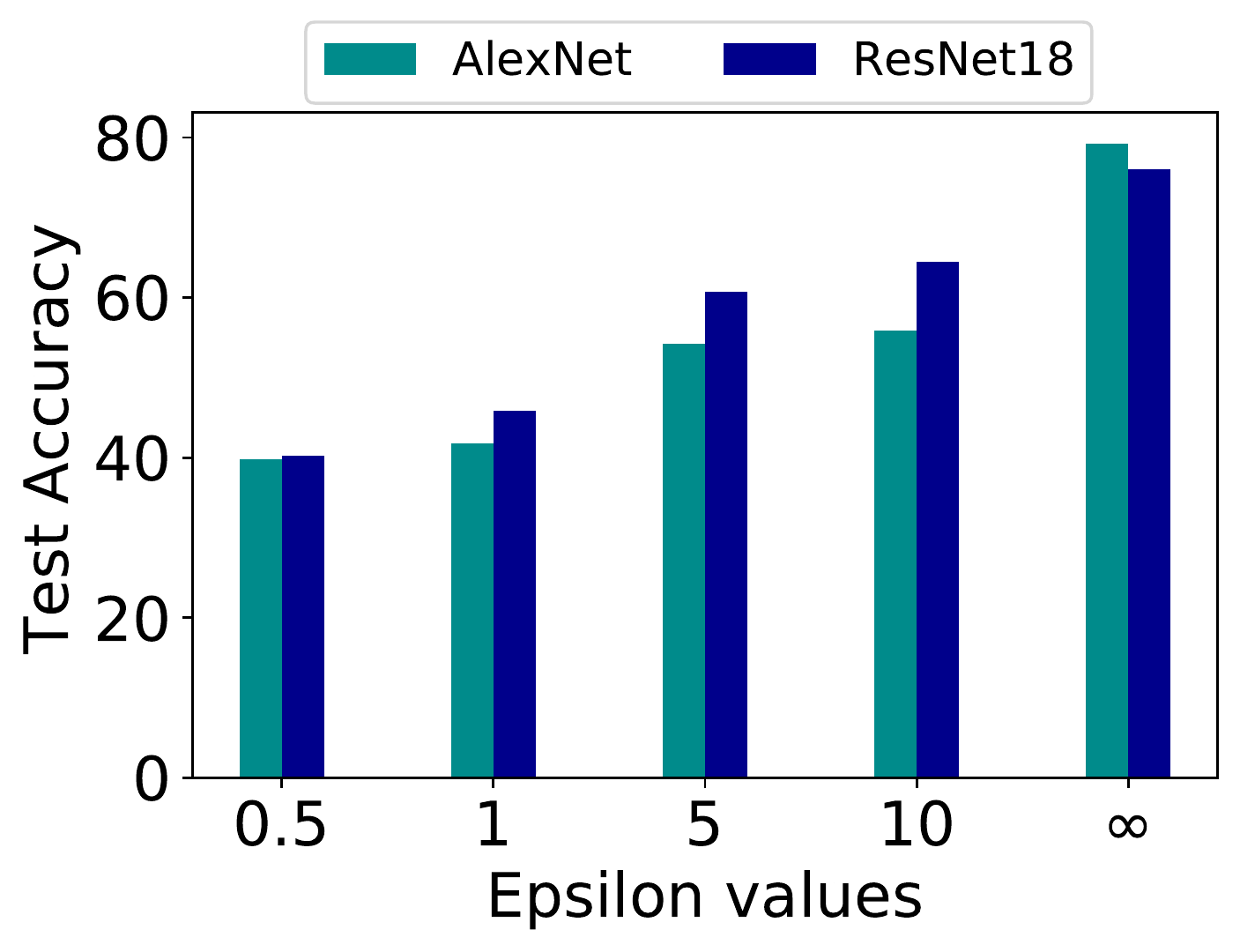}}
\caption{Comparison of Test accuracies on APTOS DP-SGD with ResNet18 and AlexNet.}
\label{aptos_dpsgd}
\end{center}
\vskip -0.2in
\end{figure}

In Fig~\ref{fig:catdogexplanation}, we see that the significant visual difference in explanation arises from $\epsilon = 1$ to $\epsilon = 5$ in both~\ref{fig:gauss_dpsgd} and~\ref{fig:laplace_dpsgd} for both the classes (``cat'' and ``dog''). Fig~\ref{aptos_dpsgd} showcases DP-SGD test accuracy results on the APTOS dataset. We notice a substantial increase in a private model ($\epsilon = 10$) when compared to a non-private model ($\epsilon = \infty$).

% \begin{table*}[h]
% \centering
% \begin{tabular}{llll}
% \hline
% Metrics      & $\epsilon = 0.25$ & $\epsilon = 0.5$ & $\epsilon = 1$ \\ \hline
% Average Drop \% & 44.13             & 15.21            & 10.57          \\
% Average Inc \%  & 35.94             & 74.89            & 82.52          \\ \hline
% \end{tabular}
% \caption{Results on Resnet-18 network with Local DP on the APTOS dataset.}
% \end{table*}

% \begin{table*}[h]
% \centering
% \begin{tabular}{llll}
% \hline
% Metrics      & $\epsilon = 0.25$ & $\epsilon = 0.5$ & $\epsilon = 1$ \\ \hline
% Average Drop \% & 36.44             & 45.64            & 19.51          \\
% Average Inc \% & 47.93             & 38.95            & 72.15          \\ \hline
% \end{tabular}
% \caption{Results on the AlexNet network with Local DP on the APTOS dataset.}
% \end{table*}

% We also introduce a simple metric to quantify the leakage in Interpretability of Differentially Private models. 

% \begin{equation}
%     \sum ^{N}_{i=1}\dfrac{f'\left( X_{i}\right)}{f\left( X_{i}\right)} \times 100
% \end{equation}
% where $f'(X)$ refers to an input image $X$ being evaluated on an $\epsilon$-DP private model $f$. For simplicity, $f(X)$ is taken as the baseline (non-DP or $\infty$-DP) model. This metric allows us to compare explanations across different privacy level models against the $\epsilon = \infty$ model, which always gives us 100\% with this metric.     

\section{Conclusion \& Future Work}

In this work, we show how privacy could affect the overall interpretability in ML models. We  quantitatively display results on different DP variants (local and global DP) on various privacy regimes. Our results (Figs~\ref{fig:ldp_aptos}, \ref{fig:avgdropsandinc_ldp}, \ref{fig:avgacc_inpmasks}, \ref{fig:catdogexplanation}) portray promising directions in this area as transparency and accountability should complement ML models in order to understand how noise affects the training of DP-trained models. We observe that there are desirable points in the three dimensional trade-off space between privacy, utility and interpretablity, where all the three metrics can have meaningful values.
%On the whole, DP-trained models allow more flexibility than standard models due to the privacy factor ($\varepsilon$) and could be commercially viable for medical imaging tasks. 
%Interpretability of diffentially private trained models is specifically necessary in medical settings in order to help understand the functions and decision making of the network.

As future work, we hope to find better heuristics for navigating this trade-off space and  devise a framework for interpretability, catered to the characteristics of DP-trained models.

\bibliography{main}
\bibliographystyle{icml2021}

\end{document}